\pdfoutput=1 

\documentclass[11pt,a4paper]{article}
\usepackage{times,latexsym}
\usepackage{url}
\usepackage[T1]{fontenc}
\usepackage{diagbox}
\usepackage[utf8]{inputenc}
\usepackage{microtype}
\usepackage{amsmath}
\usepackage{graphicx}
\usepackage[colorlinks=true, allcolors=blue]{hyperref}
\usepackage{multirow}
\usepackage{booktabs}
\usepackage{subfig}
\usepackage{makecell}
\usepackage{enumitem}
\usepackage{hyperref}
\usepackage{xr}
\usepackage{amsthm}
\usepackage{ulem}

\usepackage{tabularx}

\usepackage[acceptedWithA]{tacl2021v1}

\usepackage{xspace,mfirstuc,tabulary}

\newif\iftaclinstructions
\taclinstructionsfalse 
\iftaclinstructions
\renewcommand{\confidential}{}
\renewcommand{\anonsubtext}{(No author info supplied here, for consistency with
TACL-submission anonymization requirements)}
\newcommand{\instr}
\fi

\iftaclpubformat 

\else

\fi

\title{Probing Omissions and Distortions in Transformer-based RDF-to-Text Models}

\author{
  Juliette Faille
  \\
  CNRS/LORIA
  \\
  Université de Lorraine
  \\
  \texttt{juliette.faille@loria.fr}
  \And
  Albert Gatt 
  \\
  Utrecht University
   \\
   \texttt{a.gatt@uu.nl}
  \And
  Claire Gardent 
  \\
  CNRS/LORIA
  \\
   \texttt{claire.gardent@loria.fr}
}

\date{}

\begin{document}
  \maketitle
\begin{abstract}
  In Natural Language Generation (NLG), important information is sometimes omitted in the output text. To better understand and analyse how this type of mistake arises, we focus on RDF-to-Text generation and
  explore two methods of probing omissions in the encoder output of BART \citep{lewis-etal-2020-bart} and of T5 \citep{raffel2019exploring}: (i) a novel  parameter-free probing method based on the computation of cosine similarity 
  between embeddings of RDF graphs and of RDF graphs in which we removed some entities and (ii) a parametric probe which performs binary classification on the encoder embeddings to detect omitted entities. We also  extend our analysis 
  to distorted entities, i.e. entities that are not fully correctly mentioned in the generated text (e.g. misspelling of entity, wrong units of measurement). We found that both omitted and distorted entities can be probed in the encoder's output embeddings. This suggests that the encoder emits a weaker signal for these entities and therefore is responsible for some loss of information. This also shows that probing methods can be used to detect mistakes in the output of NLG models. 
\end{abstract}

\section{Introduction}

Neural models have drastically advanced the state of the art in Natural Language Generation (NLG) \cite{kale-rastogi-2020-text,agarwal-etal-2020-machine,guo-etal-2020-cyclegt} but are susceptible to two types of problems: \textit{omissions}, where the generated text fails to mention important information present in the input; and \textit{hallucinations}, where the generated text includes information not licensed by the input.
While various metrics \cite{balakrishnan-etal-2019-constrained,dhingra-etal-2019-handling,durmus-etal-2020-feqa,filippova-2020-controlled} and mitigation methods \cite{shuster-etal-2021-retrieval-augmentation,DBLP:journals/corr/abs-2102-02810,fan-etal-2019-using,wang-etal-2021-sketch} have been proposed to address these shortcomings, little has been done to analyze where these failures stem from. 
Furthermore, 
little work has studied omissions 
even though, just like hallucinations, these are a crucial element to mitigate in order to obtain reliable and trustworthy NLG models.

In this paper, we focus on omissions in KG-to-Text Generation, the task of generating natural language text from a Knowledge Graph~\cite{gardent-etal-2017-webnlg,castro-ferreira-etal-2020-2020,ribeiro-etal-2020-modeling}. We consider a KG consisting of Resource Description Framework (RDF) triples. Each graph is composed of a set of (subject, property, object) triples, where subjects and objects are referred to as \textit{entities}.  We study the omissions of entities in texts generated from RDF inputs whereby both the RDF data and the generated texts are in English. An illustrative example is given in Figure \ref{fig:examples_texts_omissions_distortions} which shows an example input graph, a text generated from that graph and a list of omitted 
entities. We also consider a second phenomenon, closely related to omission, which we term \textit{distortion}, where an entity is mentioned in the generated text, but its name is partially incorrect, due to misspelling of a proper name for example, or due to an incorrect number (examples are given in Figure \ref{fig:examples_texts_omissions_distortions} and in Appendix \ref{sec:appendix_annotation_instructions}, Table \ref{fig:examples_distortions}).

\begin{figure*}[!htbp]

    \small
    \centering
    \begin{tabularx}{\textwidth}{XX}
    \toprule

\textbf{RDF Graph}\\ 
\begin{tabular}{lcl}
        Nurhan\_Atasoy & award & State\_Award\_for\_Superior\_Achievement \\
        Istanbul & populationMetroDensity & 2691.0 \\
        Nurhan\_Atasoy & residence & Turkey \\
        Nurhan\_Atasoy & birthPlace & Reşadiye \\
        Nurhan\_Atasoy & residence & Istanbul \\
        \end{tabular}       \\ 
\textbf{Linearized Input} Nurhan\_AtasoyawardState\_Award\_for\_Superior\_AchievementIstanbulpopulationMetroDensity\\2691.0  Nurhan\_AtasoyresidenceTurkeyNurhan\_AtasoybirthPlaceReşadiyeNurhan\_AtasoyresidenceIstanbul \\
\textbf{Generated Text }
 Reşadiye born and Istanbul based, \underline{Guran Ataturk}, won the \underline{State Award for Superior } \underline{ Excellence}. Istanbul has a population density of 2691.0. \\
\textbf{RDF entities }
Nurhan\_Atasoy, State\_Award\_for\_Superior\_Achievement, Istanbul, 2691.0, Turkey, Reşadiye \\
\textbf{Manually annotated omissions } 
Turkey\\
\textbf{Manually annotated distortions (underlined) } 
Nurhan\_Atasoy, State\_Award\_for\_Superior\_Achievement \\
\textbf{Automatically detected omissions or distortions  } 
Nurhan\_Atasoy, Turkey\\\bottomrule

    \end{tabularx}

\caption{Example of an RDF input and Generated Text with corresponding results of the automatic entity detection, and manual annotations of omissions and distortions
}
\label{fig:examples_texts_omissions_distortions}

\end{figure*}

Our research goal is to understand where these omissions and distortions come from and whether we can predict them from the embeddings of the encoder. Specifically, 
we hypothesise that
  the encoding of an input graph whose corresponding output text involves an omission should be distinguishable from the encoding of a graph which does not lead to an omission.

A common methodology to help understand which part of a neural model is responsible for a given behaviour, is to use a probe, 
i.e., to test whether this behaviour can be predicted from the model internals.
We introduce two probes for the analysis of omissions and distortions: (i) a parameter-free probe where omissions can be detected from the model-internal representations without needing to learn new parameters; and (ii) a probing classifier trained on a dataset of KG representations which are categorised as 0 if the generated text fails to correctly verbalise some input entity and 1 if all entities are correctly verbalised. Both probes indicate that the encoder representations of  graphs which lead to an omission in the generated text can be distinguished from those that do not. To our knowledge, this is the first systematic demonstration that, in encoder-decoder architectures for 
RDF-to-text generation, the encoder plays a role in determining whether content is omitted in the output. We also provide evidence that omissions cannot entirely be predicted 
as a result of the decoding strategy used during generation and examine the features characteristic of omitted entities.
We make the following contributions.

We create two datasets of (RDF graph, generated text) pairs annotated for omissions and distortions: a dataset of 72k instances where omissions were annotated automatically and a dataset of 12k instances where the annotation was manual. We make these datasets publicly available together with the models and scripts necessary to replicate our results\footnote{\url{https://gitlab.nl4xai.eu/juliette.faille/probing-omissions}}.

We propose two main methods to detect omissions and distortions in the encoder's embeddings:
(i) a parameter-free method using 
cosine similarity between the embedding of an RDF graph and the embedding of the same RDF graph from  which we removed mentioned, omitted or distorted entities and 
(ii) a parametric method using a neural binary classifier.

We find that in most entries of our dataset, omissions and distortions can be detected in the encoder outputs by both probing methods. Using logistic regression, we further explore whether the likelihood of an entity being omitted or distorted can be predicted from its features.

The paper is structured as follows. In Section \ref{sec:method}, we describe the generation models (BART and T5)\footnote{We carry out most of the experiments on BART and do a shorter experiment on T5.} and present the (RDF,Text) data we use to fine-tune them.  We further explain how we annotate the output of these models with the (possibly empty) set of RDF entities
which are omitted or distorted in the output text.
In Section \ref{sec:analysis_omissions}, we give some motivation for the choice of our two probing methods. 
Section \ref{sec:cosine_similarities} introduces  the new parameter-free probing method we propose and presents its results. In Section \ref{sec:probing}, we present the second probe, a neural classifier, and we discuss the results obtained. We also describe different control and upper bound experiments which we use to evaluate the quality of that probe. In Section \ref{sec:logistic_regression}, we present the results of a logistic regression classifier, which investigates whether some of the dataset features can be correlated with omitted and distorted entities. 
Section \ref{sec:results_t5} summarizes the probing experiments on T5 and Section \ref{sec:limitations} outlines the main limitations of our approach.

\section{Related Work}

\label{sec:related_work}
We  first discuss work on probing pre-trained models. We then go on to discuss previous work focusing on content-related issues in NLG. 

\paragraph{Probing of Pre-trained Models}
\label{sub-sec:related_work_analysis_nlp_models}

Probing the information encoded in pre-trained models is an
active research field in NLP. However, most previous work has focused on analysing Natural Language Understanding (NLU) rather than NLG models.
In particular, \citet{belinkov-glass-2019-analysis} survey approaches which analyse NLU models including in particular, approaches based on probing classifiers.  
Other work has focused on analysing the internals of BERT. In particular, \citet{rogers-etal-2020-primer}  provides an extensive survey of the different studies which have looked at the knowledge encoded in BERT weights.   
There are also multiple papers \citep{conneau-etal-2018-cram, adi2017finegrained, tenney2019learn, ettinger-2020-bert} which  probe  the  linguistic properties encoded in the  embeddings of various NLU models (e.g. BiLSTM, Gated Conv Net), or in the different layers of the encoder. For instance, \citet{koto-etal-2021-discourse} probe the different layers of seven PLMs (including BART and T5) using tasks related to discourse structure while \citet{tenney-etal-2019-bert} study how the information from the NLP pipeline can be localised in different layers of the BERT encoder finding that low-level information appears in the first layers, whereas high-level information can be found in higher layers. 

We extend this work by analysing the source of omissions and distortions in RDF-to-Text models.
\begin{table*}[!htbp]
\centering\small
\begin{tabular}{lr|rr|rr}
  \toprule
& All  & \multicolumn{2}{c}{In Domain Graphs} & \multicolumn{2}{c}{OOD Graphs}  \\
  &   & WNLG Train & WNLG Dev & WNLG Test &  KELM \\
  \midrule
  \bf Automatic annotation &&&&&\\
Nb of texts (\%)   &71,644 (100\%) & 35,373        & 4,534       & 6,367        & 25,370 \\
Nb of texts with O & 33,160 \footnotesize{(46\%)}& 5,526 \footnotesize{(16\%)}        & 661 \footnotesize{(15\%)}       & 3,440 \footnotesize{(54\%)}       & 23,533 \footnotesize{(93\%)}\\
Nb of Os                               & 74,462 & 10,006        & 1,110       & 5,691        & 57,655 \\
                                                        &       &              &            &             &       \\
\bf Manual annotation    &&&&&       \\
Nb of texts (\% )                                   & 12,886 (18\%)& 5,526         & 661        & 3,440        & 3,259  \\
Nb of texts with O             & 6,249  \footnotesize{(48\%)} & 1,285  \footnotesize{(23\%)}         & 170  \footnotesize{(26\%)}        & 2,146  \footnotesize{(62\%)}        & 2,648  \footnotesize{(81\%)}  \\
Nb of Os                               & 9,096  & 1,374         & 176        & 3,087        & 4,459  \\
Nb of texts with D &6,518  \footnotesize{(51\%)} &  699  \footnotesize{(13\%)}& 96  \footnotesize{(15\%)} & 2,556  \footnotesize{(74\%)} & 3,167  \footnotesize{(97\%)}  \\
Nb of Ds              & 9,034 & 734 & 104 & 4,746 & 3,450 \\
Nb of texts with O or D & 8,508  \footnotesize{(66\%)} & 1,892  \footnotesize{(34\%)}         & 255  \footnotesize{(39\%)}        & 3,194  \footnotesize{(93\%)}        & 3,167  \footnotesize{(97\%)}  \\
Nb of Os and Ds               & 18,130 & 2,108         & 280        & 7,833        & 7,909 \\
\bottomrule

\end{tabular}
\caption{\textbf{Omission and Distortion Statistics for the texts generated by the BART RDF-to-Text Model} (O: ommissions, D: Distortion, WNLG: WebNLG). 
Unsurprisingly, omissions are more numerous on the data not seen at training time (OOD Graphs)}
\label{tab:figures_dataset}
\vspace{0cm}
\end{table*}
Previous work on probing has also pointed out some important methodological issues which arise when using a probing classifier. \citet{10.1162/coli_a_00422} stresses the importance of using controls and upper bounds while \citet{adi2017finegrained} criticize the lack of connection between probing tasks and downstream tasks in approaches where the embeddings of the encoder are analysed independently of any task. In line with this work, we use probing to analyse omissions in the downstream task of generation from RDF data; and we situate the results of our parametric probe with respect to both an upper bound and several control tasks. 

\paragraph{Content-related issues in text generation}
\label{sub-sec:related_work_content_issues_NLG}

Evaluating and minimizing semantic errors in the output of Neural NLG models has been the topic of extensive research. 
\citet{li2022faithfulness}'s survey paper summarizes the different evaluation and mitigation techniques which can be used to address faithfulness in NLG. 
Evaluation 
has given rise to several recent shared tasks such as \citet{gehrmann-etal-2021-gem} and \citet{thomson-reiter-2021-generation}. Multiple papers try to improve the reporting of models' mistakes by giving guidelines to avoid under-reporting of errors \citep{van-miltenburg-etal-2021-underreporting} or to provide standardized definitions of model errors \citep{howcroft-etal-2020-twenty, belz-etal-2020-disentangling}. Other work has developed metrics to measure semantic adequacy between output text and source information in summarisation~\citep{maynez-etal-2020-faithfulness, scialom-etal-2021-questeval}. 

The specific problem of hallucinations has been studied in multiple NLG tasks as shown by \citep{ji2022survey}'s survey and some work has analysed hallucinations in  Machine Translation \citep{lee2018hallucinations, raunak-etal-2021-curious}.

Overall however, relatively little work has focused on analysing errors in NLG models and to our knowledge, no work has been done so far on analysing the source of  omissions and distortions in  RDF-to-Text generation models.

\section{NLG Models and Annotated Data}
\label{sec:method}

To create the data necessary for analysing omissions in RDF-to-Text generation models, we train two RDF-to-Text Generation models and use these models to generate two large datasets of (RDF graph, text) pairs. Using both automatic and manual annotation, we then annotate the generated texts  for omissions and distortions. 

\subsection{Generation Model }
To assess generalisation, 
we train two RDF-to-Text generation models by fine-tuning BART~\citep{lewis-etal-2020-bart} and T5
~\citep{raffel2019exploring}, on the WebNLG training set \citep{castro-ferreira-etal-2020-2020}\footnote{To fine-tune the model, we linearised the RDF triplesets without adding any special tokens to separate the triples or the entities or properties within a triple.}, a dataset of 47k (RDF graph, text) pairs where the RDF graphs are subgraphs of DBPedia and texts are crowd-sourced.
We choose the WebNLG dataset as this data is a reference dataset for the RDF-to-Text task, annotated manually and therefore less likely to contain noise than synthetic or automatically scraped data. 
The absence of noise is important as we want to analyse mistakes which can be attributed to the generation model instead of mistakes coming from training on a noisy dataset.
Likewise, BART and T5 are well-suited for our experiments since they achieve state-of-the-art performance on many generation tasks, and the best-performing models in the 2020 edition of the WebNLG shared task were based on these models \citep{yang-etal-2020-improving-text, castro-ferreira-etal-2020-2020}.  The details of the fine-tuning and evaluation of the fine-tuned BART and T5 models are given in Appendix \ref{appendix:bart_details}.

As manual annotation is costly, we only perform manual annotation on the BART generated texts and therefore focus the presentation on our experiments with BART (Sections \ref{sec:method}-\ref{sec:probing}).
The additional T5 experiments 
are discussed in Section~ \ref{sec:results_t5}.

\subsection{(RDF,Text) Data}
As input to generation, we use graphs from both WebNLG 
and KELM \citep{agarwal-etal-2021-knowledge}, a large dataset of Wikidata graphs paired with synthetically generated texts. 
Using graphs from KELM increases the diversity of entities in our dataset. 
In addition, including graphs from both the KELM and the unseen part of the WebNLG test data permits comparing the performance of our probes on both in- (WebNLG train, dev and test seen) and out-of-domain (WebNLG test unseen, KELM) input graphs.

We use all 16,657 RDF graphs from the WebNLG V3.0 dataset and 6k graphs from the KELM dataset (1k graphs for each graph size from 1 to 6 triples), which yields a total of 
22,657 RDF input graphs. 

We then augment this dataset by permuting the linearised graphs as we observe that such permutations can lead to differences in the generated texts. 
The permutations we use are random changes in the order of the triples in an input graph. We randomly choose a maximum of 6 permutations of the original linearised graph. This means that for triplesets of length 2 or 3, we keep all possible permutations. For triplesets of length 4 to 7, we randomly choose 6~\footnote{In case some of the randomly chosen permutations are identical, we remove duplicates.}. This yields a total of 93,488 RDF graphs.

We then generate texts from these graphs using our BART model.
However, as not all permutations of the input graphs lead to a
different text, we keep only the 71,644 (graph,text) pairs which have
distinct output texts.

\subsection{ Annotated Data}
\label{subsec:annotated_data}
We annotate each (RDF graph, Generated Text) pair with the (possibly empty) set of RDF entities which are omitted or distorted in the generated text. 
We create these annotations 
using both automatic and manual annotation to assess  whether our probing methods yield substantially different results when relying on automatic annotations. Finding that there is little or no difference between the two methods would indicate that a cheaper, albeit noisier, automatic method can still yield useful insights.

For the automatic detection of omissions, we use the algorithm developed in \citet{faille-etal-2021-entity-based}, which has a recall of 0.74 and a precision of 0.75 on WebNLG data.

For the manual annotation, we asked three annotators to check the automatic detection
focusing on texts for which at least one omission was found.\footnote{Note, that this potentially introduces a bias as false negatives of the automatic detection won't be included in the manually annotated dataset. This is, however, the best solution we could find to focus the annotation work on texts with omissions and distortions and avoid spending annotation time on correct texts. }
For each entity in the input RDF graph, they annotated whether the entity is mentioned in the text, omitted or distorted. 
Annotators are for instance instructed to label as distortions misspelling of proper names, partial omission of entities or wrong units of measurement. 
The three annotators were  students following an English taught Master in NLP who had native (2) or high proficiency (1) in English. There were employed by the University on a 20 hour/month work contract with a salary determined by  the University grids and including  social benefits in line with the host country work laws. The annotators instructions are given in Appendix \ref{sec:appendix_annotation_instructions}. 

We computed the Cohen's kappa score between each pair of annotators on a subset of 100 texts (20 instances randomly chosen from the four dataset subsets, WebNLG train, dev, test and KELM). The scores vary from 0.56 to 0.69, which is considered to be substantial agreement \citep{artstein-poesio-2008-survey}.

All texts (71,644 instances)  were automatically annotated for omission and 12,886 of these were manually annotated for both omission and distortion. 
Table \ref{tab:figures_dataset} reports the total number of manually and automatically annotated texts for each dataset, the
number of texts which have at least one omission/distortion and the total number of omissions/distortions.

\subsection {Data for the probing experiments}
\label{subsec:data_probing}
For our probing experiments, we keep annotated texts with at least one omission and divide them into train (70\%), dev (15\%), and test (15\%) sets. 
Of the distinct entities which are omitted or distorted, approximately 50\% occur in the train split, and 25\% occur in the dev and test splits. Hence, the dev and test splits contain omissions and distortions of entities which are not seen during training.

\subsection {Evaluation of Automatic Annotation}
\label{subsec:evaluation_automatic annotation}
To assess the quality of the automatic annotations, we compare them with the manually labeled omissions\footnote{We only compare to omissions, not distortions, as \cite{faille-etal-2021-entity-based}'s detection algorithm aims to detect omissions only.}. The F-measure is 0.58 (Precision: 0.52, Recall: 0.66). We therefore cannot consider the automatic annotation as completely reliable for our probing experiments. We use it in two ways: (i) as a way to speed up the annotation process as described in subsection \ref{subsec:annotated_data} and (ii) as an addendum 
to our main experiments on the manually annotated data. 

\subsection{Exploring the possible role of decoding strategies}
\label{sec:decoding_strategies}
The hypothesis we test in this paper is that omissions and distortions are substantially
due to problems with encoding. However, it is possible that the decoding strategy used also plays a role. 
To check whether  decoding impacts omissions, we  
compare four  decoding strategies : greedy, beam search, top-k \citep{fan-etal-2018-hierarchical}  and top-p (i.e. nucleus sampling, \citep{Holtzman2020The}), with k=50 and p=0.9 (which are standard values for top-k and top-p decoding).
We generate texts from RDF graphs from the WebNLG test set (1,779 graphs) using each of these four strategies; we then use the automatic detection of omissions on the generated texts. For each text, we compute the Intersection over Union of omitted entities between decoding strategies (we report the mean and median of IoU across the texts in Table \ref{tab:intersection_decoding_strategies}, in the rows marked `Auto').
Even though some differences emerge, we find that the entities omitted by BART remain to a large extent similar across the four different decoding strategies. We also find that different decoding strategies result in very similar numbers of texts with at least one omission (844 for greedy decoding, 985 for beam search with 5 beams, 980 texts for top k with k=50 and 974 for top p with p =0.9).  

As discussed in section \ref{subsec:evaluation_automatic annotation}, the automatic annotation has two main limitations: it is noisy and it does not distinguish omissions from distortions. In order to have a more reliable and fine-grained analysis of the impact of the different decoding strategies on omissions and distortions, we manually annotated the texts generated from 500 input graphs (taken from the WebNLG test set) using the four decoding strategies, which yields a total of 2k manually annotated (graph, text) pairs. We then computed the IoU scores for omissions and distortions on this data. The results are reported in Table \ref{tab:intersection_decoding_strategies} (rows marked `Manual-O' and `Manual-D'). The results for the manually annotated data are similar to the results on automatically annotated data. The overlap of omitted entities between the different decoding strategies is quite high, with a mean IoU score across texts ranging from 0.53 to 0.63. For distorted entities, the overlap is even higher, with mean IoU between 0.69 and 0.81. This suggests that the impact of the decoding strategy on omissions is limited, and is even more limited on distortions. Nevertheless, the potentially limited role of the decoding strategy is worth investigating in more detail; we leave this as a topic for future work.

\begin{table}[!htbp]
\small
 \begin{tabular}{ll|lll}
                 & &Greedy & Beam & Top k \\\hline
      
\multirow{3}{*}{Beam}& Auto     & 0.61/1.0    &         &                     \\
& Manual-O     & 0.62/1.0    &          &                     \\
& Manual-D     & 0.81/1.0    &         &                     \\ \hline
\multirow{3}{*}{Top k}  & Auto          & 0.52/0.5    & 0.73/1.0         &                     \\
 & Manual-O          & 0.53/0.5    & 0.56/1.0         &                     \\
  & Manual-D           & 0.69/1.0    & 0.71/1.0         &                   \\ \hline
\multirow{3}{*}{Top p}& Auto  & 0.53/0.5    & 0.73/1.0 & 0.76/1.0     \\ 
& Manual-O  & 0.57/1.0    & 0.63/1.0 & 0.63/1.0     \\ 
& Manual-D  & 0.72/1.0    & 0.74/1.0 & 0.79/1.0     \\ 
\end{tabular}
\caption{Mean/Median of Intersection over Union scores for each text of omitted/distorted entities (using automatic annotation, on 7116 texts), of omitted entities (using manual annotation of 2000 texts), of distorted entities (using manual annotation of 2000 texts) for different decoding strategies }
\label{tab:intersection_decoding_strategies}
\end{table}



The fact that omissions and distortions remain roughly the same when we modify the decoding strategy suggests that the decoding strategy is probably not the main cause of omissions. We should therefore look for the cause of omissions in some other part of BART. In this paper we choose to focus on the encoder. For simplicity and as the decoding strategy does not seem to exert much impact on omissions, in what follows we only consider greedy decoding.

\section{Method}
\label{sec:analysis_omissions}

To study omissions and distortions, we choose probing rather than some other analysis methods as it provides a direct, model-agnostic and relatively computationally inexpensive way to study a model internals. Furthermore, as discussed in section \ref{sec:related_work}, probing has been shown to be effective for analysing how linguistic phenomena are handled in neural architectures. 

We design two probes to test whether omissions 
can be detected from encoder representations.
The first probe is a parameter-free probe which tests whether
the encoding of a graph leading to an omission contains less information about an omitted entity than about a mentioned entity (an entity that is mentioned in the output text).

The second probe is a standard probing classifier which seeks to distinguish the encoded representations of input graphs leading to an omission from the  representations  of input graphs whose output text correctly verbalises all input entities.

While probing classifiers have been widely used to analyse neural networks, \citet{hewitt-liang-2019-designing} observe that a probe might memorise the data it is trained on rather than evaluate the information present in the internal representations it is designed to analyse. The parameter-free method we propose in Section~\ref{sec:cosine_similarities} helps address this issue as it extracts information directly from the internal representations without learning new parameters.  For the parametric probing classifier of Section~\ref{sec:probing}, we address this issue by introducing a control task which can only be resolved by memorising the data, and  computing selectivity i.e., the difference between the probe performance on the control task and on the probing task. We show that our probing classifier has both high performance, which indicates that omissions can be detected from the encoded input graphs and high selectivity, which indicates that the probe is learning rather than memorising the data.

\section{Parameter-free Probing}
\label{sec:cosine_similarities}

This method is based on the intuition that the encoder representations of RDF graphs which lead to omission have a weak signal for the omitted entity. We hypothesise that, because it lacks specificity, the representation of an omitted entity is more similar to the  representation of the unknown token <unk> than the representation of an entity that is correctly verbalised in the output text. We test this hypothesis by comparing the similarity between $g$, the encoder representation of a graph leading to an omission with two alternative representations:
(i) $g^{\setminus o}$, the embedding of the same RDF graph in which an omitted entity $o$ has been replaced with the unknown
special symbol <unk>; and 
(ii) $g^{\setminus m}$, the embedding of this graph in which  a mentioned entity has been replaced with <unk>.
If the encoded representation of  omitted entities lacks specificity, we expect the embedding of $g$ to be closer to the encoding of $g^{\setminus o}$ (where  an omitted entity has been replaced with  <unk>) than to that of  $g^{\setminus m}$ (where a mentioned entity has been replaced with  <unk>). 

Let $G$ be the set of graphs in our data which have at least one mention and one omission.
For each graph $g\in G$ 
which is associated with $J_g$ omissions or distortions and $K_g$ mentions, we denote $\{o_j\}_{j=1,...,J_g}$ the omissions/distortions and $\{m_k\}_{k=1,...,K_g}$ the mentions in $g$.

For each graph embedding $g\in G$, we compute an average of similarities between  $g$ and  $g^{\setminus m_k}$:
$$
sim(g,g^{\setminus M}) = \frac{1}{K_g} \sum_{k=1}^{K_g} sim(g,g^{\setminus m_k})
$$
and similarly for the average of similarities between $g$ and $g^{\setminus o_k}$.
We then compute the proportion of texts 
in our dataset for which :
$$sim(g,g^{\setminus o_j}) > sim(g,g^{\setminus M})\text{ ,}$$ 
where, for two embeddings $g$ and $g'$,
$$sim(g, g')=cos(mean(g), mean(g'))$$

is the cosine similarity on fixed-sized graph embeddings obtained by mean pooling as  fixed-size embeddings are needed
to compare embeddings of graphs of arbitrary length using cosine similarity. 
Various pooling strategies can be used. For instance, \citet{reimers-gurevych-2019-sentence} experiment with three strategies: using the output of the CLS token, and max or mean pooling of all the output vectors. 
As they find that mean pooling performs best, 
we experiment with two different mean pooling strategies (dimension-based and token-based averaging) as follows.

The output of the encoder is a matrix of shape $T \times 1024$, with $T$ the number of tokens. 
As the number of tokens $T$ varies with the length of the entities and properties, we first average the representation of each entity and property and obtain a matrix of size $N \times 1024$, with $N$ the number of entities and properties in the RDF graph.
We then can average this matrix either row-wise (i.e. dimension-based averaging) and obtain a mean vector of length $1024$ 
or column-wise (i.e. token-based averaging) and obtain a mean vector of length $N$. As the two averaging strategies give similar results, for simplicity we report only the dimension-based averaging results. Note that another option would have been to compute the cosine similarities on flattened embedding matrices. It would be interesting to compare the results of the probe on  flattened and averaged matrices. We leave this as a topic for future work.

\begin{table}[!htbp]
\centering\small
\begin{tabular}{l|l|lll|ll}
\toprule
& \bf All & \multicolumn{3}{c}{\bf In Domain} & \multicolumn{2}{c}{\bf OOD}\\
& &  W-T &  W-D & W-S & W-U & K \\
\midrule
\footnotesize{Manual} &&&&&&\\
O & 0.68 & 0.64 & 0.72 & 0.61 & \textcolor{gray}{\textit{0.52 }}   & 0.77    \\
D & 0.44 &  0.70 & 0.68&  0.47 & 0.45 & 0.47    \\
\footnotesize{Auto}  & \footnotesize{0.66} & \footnotesize{0.83} & \footnotesize{0.85} & \footnotesize{0.56} & \footnotesize{0.44} & \footnotesize{0.65}    \\
\bottomrule
\end{tabular}
\caption{\textbf{Proportion of graphs for which $sim(g, g^{\setminus o_j}) > sim(g, g^{\setminus M}) $}, O: Omissions, D: Distortion, W: WebNLG, T/D/S/U/K: Training/Development/Seen/Unseen/Kelm Data.
  The figures in gray correspond to non-statistically significant results. A proportion of 50\% indicates failure to distinguish omissions/distortions from mentions. Proportions larger (resp. smaller) than 50\% indicate that omissions/distortions can be distinguished from mentions and that our assumption that encodings leading to an omission/distortion have a weaker signal than the ones leading to a mention is supported (resp. contradicted). 
}
\label{fig:results_cosine_similarities}
\end{table}

\paragraph{Results}
\label{subsection:similarity_results}
Table \ref{fig:results_cosine_similarities} shows  the proportion of examples in which  $sim(g, g^{\setminus o_j})$ is greater than  $sim(g, g^{\setminus M})$ for the different subsets of our data.
For statistical significance testing, we used a one way chi-square goodness-of-fit test comparing the numbers of samples in each subset of the data for which $sim(g, g^{\setminus o_j}) > sim(g, g^{\setminus M}) $, to those where this inequality does not hold.
As we did multiple tests for each subset of the data, we adjusted the p-values using Bonferroni correction.
 
On average, the proportion of graphs for which $sim(g, g^{\setminus
  o_j}) > sim(g, g^{\setminus M}) $ is 66\% 
for the automatically annotated data and 68\% 
for the
manually annotated data. Most results are statistically significant 
which supports our hypothesis that the encoding of a graph leading to
an omission has weak signal for that entity.

The subsets of the data on which this proportion is the lowest and the results are not statistically significant are the subsets of the data unseen during training, the WebNLG Test Set and KELM. This trend is even clearer on the subset of the WebNLG test set with entities and (DBPedia) categories unseen during the model's fine-tuning (W-U in Table \ref{fig:results_cosine_similarities}). 
We hypothesise that this is because graphs which are different from the graphs seen during the fine-tuning of the generation model are less distinctively encoded that the ones seen during fine-tuning.

The results also show a clear difference between distortions and omissions.
This suggests that the encodings of distortions and omissions are qualitatively different 
and that distortions are not merely a type of omission.

 Finally, we observe that while there is a difference between results on the automatically and manually annotated data, the delta is relatively small (0.66 on avg for the Auto data vs. 0.68 for the manual data),  and the overall trend is the same -- which indicates that \cite{faille-etal-2021-entity-based}'s omission detection algorithm performs well enough on out-of-domain data to be used for such experiments.

\textbf{In this section, we showed that a parameter-free probing method, based on cosine similarities, can be used to probe omissions and distortions in the encodings of RDF graphs.}

\section{Parametric Probing: Binary classifiers}
\label{sec:probing}
Our second probe is a binary classifier which takes as input  the encoder representation of an RDF graph $g$ and of an entity $e\in g$, returning 0 if $e$ is omitted or distorted in the output text, and $1$ otherwise. This can be viewed as modeling an entailment relation between a graph representation and an entity: the classifier returns 1 if the graph entails the entity (mention) and 0 if it does not (omission or distortion). 

In this section we investigate whether neural binary classifiers can be used to detect omissions and distortions from RDF graph encodings.
In subsections \ref{subsec:probes_characteristics} and \ref{subsec:evaluation_metrics_probes}, we present our probes and the evaluation metrics we use. In subsections \ref{subsec:upper_bound} and \ref{subsec:controls}, we introduce an upper bound and two control tasks which help get better insight into the maximal performance our probes could reach, as well as  address the risk that our probes are not really using the information in the embeddings, but instead rely on patterns linked to the entities' identity. In subsection \ref{subsec:results_probing_classifiers}, we discuss and compare the results of the probe, of the upper bound and of the control probe.
 We also report results on hard examples (entities that can be both mentioned, omitted, or distorted) and provide additional evidence that the probing results do not rely exclusively on entities' identity. 
In addition, we include two subsidiary experiments: one aiming at comparing the results of the probes trained on the automatic versus the manually annotated dataset and another (Appendix \ref{appendix:omissions_distortions}) which  investigates the difference between omissions and distortions. The results show that probes trained on detecting omissions do not generalize well to distortions (and vice versa).

\subsection{Models} 
\label{subsec:probes_characteristics}
Our second probe is a neural classifier with fully-connected linear layers and sigmoid activation functions trained using a cross-entropy loss and the AdamW optimizer. It takes as input the average vector of the embedding matrix (dimension-based averaging, as in Section~\ref{sec:cosine_similarities}). We experiment with both a  single- (N1) and a two-layer (N2) network, using the train and test sets defined in \ref{subsec:data_probing}. For each  network, we train a model and perform hyper parameter tuning for each dataset (Manual-O+D, Manual-O, Manual-D and also Auto)\footnote{We also experimented with SVM and Decision Tree classifiers, which performed slightly worse than neural ones. For brevity, we report only the results of the neural classifiers.}.
Experimental details  are given in Appendix \ref{sec:appendix_hyperparameter_tuning}. 

\subsection{Evaluation metrics}
\label{subsec:evaluation_metrics_probes}
Our data is unbalanced as it contains more mentioned entities than omitted or distorted ones. Since we are primarily interested in detecting omissions or distortions, which we define as class 0 in our binary classification, we evaluate the results of our probes using the F-measure for class 0. We also report the balanced accuracy (B.Acc), defined as:
$$\frac{1}{2} \left( \frac{TP}{P} + \frac{TN}{N}\right) \text{ ,}$$ where $TP$ is the number of true positives, $P$, the number of positive examples, $TN$ the number of true negatives and $N$, the number of negative examples.

As suggested in \citet{10.1162/coli_a_00422}, we compare our probes to both an upper bound and a control task.

\subsection{Upper Bound}
\label{subsec:upper_bound}
Entities that are not present in the input graph can be seen as an extreme case of omission. 
Therefore a natural upper bound for our probing classifier 
is a classifier which seeks to distinguish mentioned entities from entities not present in the input graph. We train such a classifier on 18k RDF graphs (randomly selected from our dataset) and 198k entities (which are either mentioned in the RDF graph or randomly selected from the set of all entities in the dataset and that are not in the input RDF graph). Note that the training data for this classifier is different from that used for our omission probes.

The classification results for this upper bound, which are reported in Table \ref{tab:f-measures_probing_omissions_vs_mentions} (last two lines), are high
for both N1  (F1:0.91) and N2 (F1:0.97),  
showing that it is possible to detect whether or not an entity is present in the embedding of an RDF graph. Furthermore, the comparatively  lower scores obtained when probing for omissions (F1: 0.69) and distortions (F1: 0.79) suggests that
omissions and distortions are harder to detect than entities not present in the input graph. 

\subsection{Control tasks}
\label{subsec:controls}

We implement two control tasks to verify that our probes are predicting omissions from graph embeddings rather than memorising the training data.

\subparagraph{Training on randomized probing labels}
Following \citet{10.1162/coli_a_00422}, we randomize the labels in the training set and compute the F-measure and Weighted Accuracy of a control probe, trained on the randomised dataset. Table \ref{tab:f-measures_probing_omissions_vs_mentions} shows that there is a clear drop in performance when training the probes with randomised labels. We can therefore conclude that our probes are not just memorising the training set.

\subparagraph{Testing with randomized NLG model encoder}
The second control task consists in randomizing the weights of the encoder.
We randomly initialize the weights of the BART encoder and train and test the neural probe with two linear layers (our best probe) on the manually annotated dataset for omissions. We try 5 different random seeds and obtain a mean B.Acc of 0.49 (with standard deviation 0.02) and mean F-measure for class 0 of 0.31 (with standard deviation 0.13). This significant performance drop  shows that the probing results are not based on mere word identity (which could be captured by random encodings). 

\subsection{Results}
\label{subsec:results_probing_classifiers}

\begin{table}[!htbp]
\centering
\small

\begin{tabular}{lll}
  \toprule
  & N1 & N2 \\
  \midrule
  \bf Manual-O+D &\\
  F1 (B.Acc)    & 0.73 (0.78) &\bf 0.82 (0.85)\\
  C$_{F1}$ (C$_{B.Acc}$)  & 0.00 (0.50)   & \bf 0.00 (0.50) \\
  \bf Manual-O &\\
  F1 (B.Acc)   & 0.60 (0.72) & \textbf{0.69} (0.77)
  \\
  C$_{F1}$ (C$_{B.Acc}$)  & 0.00 (0.50)   & \bf 0.00 (0.50) \\
  \bf Manual-D &\\
  F1 (B.Acc)    & 0.71 (\textbf{0.88}) & \textbf{0.79} (0.79)
  \\
  C$_{F1}$ (C$_{B.Acc}$) & 0.00 (0.50)   & \bf 0.00 (0.50) \\
  \midrule
  \bf \footnotesize Auto & \\
  \footnotesize F1 (B.Acc)  & \footnotesize 0.60 (0.71)   & \footnotesize \bf 0.79 (0.83) \\
  \footnotesize C$_{F1}$ (C$_{B.Acc}$)  & \footnotesize 0.00 (0.5)   & \footnotesize \bf 0.00 (0.50) \\\midrule
  
  \bf Upper-Bound &\\
  F1 (B.Acc)   & 0.91 (0.92)& \textbf{0.97 (0.98)}
  \\
  
\bottomrule  

\end{tabular}
\captionsetup{justification=justified, width=\columnwidth} 
\caption{F-measure of class 0 (F1), Balanced Accuracy (B.Acc) for each probe and its control C$_{F1}$and C$_{B.Acc}$
.
N1/N2: One/Two Layer Network (with hyperparameters tuned on Manual-O, Manual-D or Manual-O+D). In bold, the best results for each dataset. 
} \label{tab:f-measures_probing_omissions_vs_mentions}

\end{table}

\paragraph{Best Probes}
Table~\ref{tab:f-measures_probing_omissions_vs_mentions} shows the F-measure scores and balanced accuracies for the various probes on the different datasets.
 Overall, we find (i) that the two-layer neural network performs best  and (ii) that models trained on manually annotated data perform better than those trained on automatically annotated data.

\begin{table}[!htbp]
\centering\small
\begin{tabular}{l|l|lll|ll}
\toprule
& \bf All & \multicolumn{3}{c}{\bf In Domain} & \multicolumn{2}{c}{\bf OOD}\\
 & &  W-T &  W-D & W-S & W-U & K \\
\midrule
\bf O+D&&&&&&\\
$\;$F1 &\bf 0.82 & 0.71& 0.64 &  \bf 0.85&  \bf 0.86 &  \bf  0.82
\\
$\;$B.Acc & 0.85  & 0.80  &  0.78 &  0.87   &  0.97  &  0.83 \\
\bf O &&&&&&\\
$\;$F1       & 0.69 & 0.57 & 0.57 & 0.71 & 0.69       &  0.73\\
$\;$B.Acc  & 0.77    & 0.71 & 0.71   & 0.78   & 0.77   & 0.79\\
\bf  D&&&&&&\\
$\;$F1 & 0.79 &  \bf 0.80 &  0.83 & 0.82& 0.79  & 0.78 \\
$\;$ B.Acc& 0.84  & 0.85   & \bf 0.88   & 0.86    &  0.83   & 0.83 \\
\bottomrule
\end{tabular}
\caption{F-measure of class 0 (F1) and Balanced Accuracy (B.Acc) on different subsets of the probing test set using N2. 
In bold, the best results for each dataset.
}
   \label{tab:probing_results_per_subset}
\end{table}

\paragraph{Detecting Omissions and Distortions }
Focusing on the subset of the data that was manually annotated (Manual-X, Table
\ref{tab:f-measures_probing_omissions_vs_mentions}), 
 we find that the parametric probes successfully classify  omissions (F1: 0.60, 0.69) and distortions (F1: 0.71, 0.79) 
 indicating a significant difference between the embeddings of mentioned entities on the one hand and the embeddings of distorted/omitted entities on the other hand. 


Table \ref{tab:probing_results_per_subset} gives a more detailed picture of the results for the best probe (N2) on the different subsets of the manually annotated dataset. For statistical significance testing, we used a chi-square test of independence.
Once again, we use a Bonferroni correction for the p-values, since we did a test for each of the test data subsets. All results are statistically significant. For most subsets, we observe a similar trend as on the full datasets (Table \ref{tab:f-measures_probing_omissions_vs_mentions}): probing for omissions and distortions together (Manual-O+D) yields the best F-measure on class 0 (omission and distortion) and probing for distortions (Manual-D) yields better results than probing for omissions (Manual-O). These results are different from the ones obtained with the parameter-free probe, where omissions were easier to probe than distortions. This can be explained by differences between the two probing methods:  
the parameter-free probing considers only differences of cosine similarities between embeddings, while the neural probes look at non-linear differences between embeddings. These results suggest that the neural probes are complementary to the parameter-free probing and better suited for the analysis of distortions.

\paragraph{Testing on Hard Examples}
In different texts the same entity can be mentioned, omitted or distorted.
Such cases permit testing whether our probe accurately classifies graphs that contain omissions and distortions rather than graphs that contain specific entities.
We test our best probe (N2) on such entities and report the results in Table \ref{fig:test_on_mixed_entities}.
Comparing these results with those on all entities (Table \ref{tab:f-measures_probing_omissions_vs_mentions}), we see that both the F-measure of class 0 and the balanced accuracy are similar for all manually annotated data (and even higher on the manually annotated omissions). Only on the automatically annotated dataset are the results slightly lower. These results show that the probe also performs well on difficult examples. 

\begin{table}[!htbp]
  \centering\small
  \begin{tabular}{l|lll}
      \textbf{Training Data} & \textbf{Test Data}& \textbf{\%}& \textbf{F1 (B.Acc)} \\\hline
      Manual-O & M\&O& 13\%&0.81 (0.74)\\
      Manual-D & M\&D& 14\% & 0.84 (0.81)\\
      Manual-O+D & M\&O\&D& 9\% & 0.78 (0.82) \\
      Manual-O+D & M\&O& 13\% & 0.82 (0.82) \\
      Manual-O+D & M\&D& 14\% & 0.78 (0.81) \\
      \footnotesize Auto & \footnotesize M\&O& \footnotesize 31\% & \footnotesize 0.7 (0.63)\\
      
    \end{tabular}
  \caption{Results of the N2 probe on entities that are mentioned and omitted (M\&O), mentioned and distorted (M\&D) or mentioned, omitted and distorted (M\&O\&D). The third column gives the proportions of such entities in the manual and the automatic dataset.}
  \label{fig:test_on_mixed_entities}
\end{table}

\paragraph{Correlations between Models trained on Automatically vs. Manually Annotated Data}
We want to know to what extent the automatically annotated data, which is cheap to obtain in comparison with manually annotated data, leads to the same probing results, compared to the manually annotated data. To answer this question we compute the Spearman correlation coefficient between the labels predicted by the probes on the automatic and manual datasets. We also compute the Pearson correlation coefficient between the probabilities of the class 0 predicted by the probe for all the test set examples on the automatic and the manual datasets. The correlation between the labels tells us how well the predicted classes are correlated and the correlation between the probabilities of the class 0 additionally gives an indication about whether the two probes have similar levels of confidence in their predictions.

The results are shown in Table \ref{fig:correlation_results}. The correlation strength between the manually labeled omissions and the automatic data both for the labels and for the probabilities of class 0 is high. However, there is no correlation for the labels or for the probabilities of class 0 for distortions identified from manual and from automatic data. This can be explained by the fact that the automatic mention detection from \citet{faille-etal-2021-entity-based} includes approximate string matching. Since approximate matching uses a threshold, a distortion which overlaps substantially with the entity would be counted as a mention and will therefore be labeled as belonging to class 1 by the probing classifier trained on automatic data, whereas it will be labeled as class 0 by the probing classifier trained on the manual dataset.

\begin{table}[!htbp]
  \centering\small
\begin{tabular}{llllll}
\toprule
 && \multicolumn{3}{c}{Correlation coefficient/P-value}\\ && O &D& O+D\\\midrule
\multirow{2}{*}{N1} & Sp & 0.70/0.00   & 0.07/0.10  & 0.55/0.00  \\
                          & Pe    & 0.77/0.00 & 0.02/0.70                     &0.54/0.00 \\
\multirow{2}{*}{N2} & Sp & 0.70/0.00   &0.07/0.10& 0.55/0.00     \\
                          & Pe & 0.99/0.00   &0.01/0.08 & 0.59/0.00    
                          \\\bottomrule
\end{tabular}
\caption{Correlation between models trained on Manual vs Automatically annotated data (Sp: Spearman on labels, Pe: Pearson on probabilities of class 0). P-values are computed against the null hypothesis that the correlation is no different from zero.}
   \label{fig:correlation_results}  
\end{table}

\section{Dataset Analysis using Logistic regression}
\label{sec:logistic_regression}

We train a logistic regression classifier to predict whether an entity will be omitted or distorted based on data-specific features (the 12 input features we use are detailed in the Appendix \ref{appendix:features_logistic_regression} and are about the tripleset, the entity itself and the frequency of the entity in the dataset). If such a classifier proves highly predictive, the feature weights can help understand which features impact the omission or distortion of an entity. 
We compute the logistic regression on Manual-O, -D and -O+D datasets, randomly choosing 90\% of the omissions or distortions as training set and the remaining 10\% as test set (as the results were unstable, we repeated the training and test with three different random seeds and report the mean). The results are shown in Table \ref{fig:logistic_regression_results}. 
The logistic regression does not perform well on the Manual-O dataset but performs better on Manual-D and Manual-O+D. This shows that whereas omissions seem difficult to characterize based on dataset features, features describing distortions are easier to find. In particular, we find that the position of the first occurrence of the entity in the tripleset, the number of occurrences in the tripleset, the semantic role in the graph (subject, object or both) and the length of the tripleset are the most relevant features for a distortion to happen.

\begin{table}[!htbp]
  \centering\small
\begin{tabular}{lll}
\toprule
 & \multicolumn{2}{c}{F1}\\ & Train& Test\\\midrule
Manual O &0.36&0.37\\
Manual D &0.60&0.59\\
Manual O+D &0.72&0.76\\
                          \bottomrule
\end{tabular}
\caption{F-measure of class 0 of the logistic regression on the different datasets }
   \label{fig:logistic_regression_results}  
\end{table}

\textbf{In this section, we showed that a logistic regression classifier trained on dataset features does not perform well on detecting omissions, suggesting that information for detecting them can only be found in more complex non-linear features. The same logistic regression classifier seems to indicate that some dataset features are correlated with distortions. }

\section{Study of a T5 encoder}
\label{sec:results_t5}
We claim that our probing methods are model-agnostic and can be used to analyse any encoder in a transformer-based NLG model. 
To support this claim and check whether our results can generalize to another encoder, we carry out experiments on T5  \cite{raffel2019exploring}. As with BART, we fine-tune T5-small on the training set of WebNLG (the fine-tuning details are given in the Appendix \ref{appendix:bart_details}). We then generate texts, filter out duplicates and automatically detect omissions as described in Section \ref{sec:method}. Table~\ref{tab:figures_dataset_t5} shows the statistics of the resulting corpus. Note that we did not create a manually annotated corpus for T5. The results are therefore limited to automatically detected omissions.
We probe the T5 encoder using the parameter-free probing method described in Section \ref{subsection:similarity_results} and the 2-layer neural probing classifier described in Section \ref{sec:probing} (we use the same hyperparameters as the probing classifier N2-O+D). The results are reported in Table \ref{fig:results_t5}. 
\\
The results for the probing classifier are similar to those obtained for the BART encoder, with an F-measure of class 0 of 0.8 and a Balanced Accuracy of 0.85 (compared to BART: F1=0.82 and BAcc=0.85). This suggests that our probing method generalises well to the T5 encoder.
\\
The results for the parameter-free probing are much better with T5 than with BART which suggests that the applicability of this parameter-free probing depends on the type of embeddings/encoder studied. This should be studied in future work to determine how the choice of the model and of the fine-tuning parameters impact the performance of this probing method. 

\textbf{In conclusion, our two probing methods can be used to analyse omissions in the embeddings of another transformer-based encoder.}

\begin{table}[!htbp]
\centering\small
\begin{tabular}{lrrrr}
  \toprule
& \# T & \# T(O) & \# O \\
  \midrule
  \bf WebNLG &&&\\
  $\;$Train & 36,704 & 7,064 \footnotesize{(19\%)} & 7,824 
  \\
  $\;$Dev &4,658& 882  \footnotesize{(19\%)}& 993              
  \\
    $\;$Test &6,173 & 2,286 \footnotesize{(37\%)} & 2,855 \\
   \bf KELM & 24,963 &17,852 \footnotesize{(72\%)} & 29,596\\ 
  \bf ALL & 72,498 & 28,084\footnotesize{(39\%)}&  41,268\\
  \bottomrule
\end{tabular}
\caption{\textbf{Corpus Statistics for texts generated by T5} (T: Texts, T(O): Texts with Omissions, O: Omissions)
}
\label{tab:figures_dataset_t5}
\end{table}

\begin{table}[!htbp]
\centering\small
\begin{tabular}{ll|lll|ll}
\toprule
 & \bf All & \multicolumn{3}{c}{\bf In Domain} & \multicolumn{2}{c}{\bf OOD}\\
  &  W-T &  W-D & W-S & W-U & K 
\\\midrule
 \bf NP.P  &&&&&\\
   \multicolumn{1}{r}{\scriptsize{T5}} & 0.89 &0.84 & 0.84  & 0.88 & 0.81  & 0.91     \\
   \multicolumn{1}{r}{\scriptsize{BART}} & 0.66 &0.83 & 0.85  & 0.56 & 0.44  & 0.65     \\
\midrule
\bf P.P &&&&&\\
\textbf{F1} &   &&&&&\\
\multicolumn{1}{r}{\scriptsize{T5}}  & 0.8   & 0.84 & 0.83 & 0.79 & 0.7 & 0.78   \\
\multicolumn{1}{r}{\scriptsize{BART}} & 0.69  & 0.57 & 0.57 & 0.71 & 0.69 & 0.73   \\
\textbf{B.Acc } &&&&&\\
 \multicolumn{1}{r}{\scriptsize{T5}} & 0.85 & 0.88 & 0.88  &0.83 & 0.77  & 0.81 \\
\multicolumn{1}{r}{\scriptsize{BART}} & 0.77 & 0.71 & 0.71  &0.78 & 0.77  & 0.79 \\
\bottomrule
\end{tabular}
\caption{\textbf{Results of parameter-free (NP.P) and parametric (P.P) probing of the T5 encoder.} We also recall the results on the BART encoder. All the results are statistically significant results (using chi-square goodness-of-fit for NP.P and independence for P.P tests with Bonferroni correction with alpha=0.05). 
NB: The results NP.P and P.P are not directly comparable, as they are based on different metrics.
}  
\label{fig:results_t5}
\end{table}

\section{Limitations and Future Work}
\label{sec:limitations}
\paragraph{NLG Task studied}
In this paper, we restrict our study to NLG models that generate English text from RDF data. In future work, our study could be extended to other Data-to-Text or Text-to-Text tasks as well as to other languages. A key bottleneck is the annotation of omissions and distortions, in particular whether it should be automatic or manual.
\paragraph{Scope of the study}
We study two state-of-the-art models for Data-to-Text, BART and T5, and probe their encoders. Further work in that direction 
could for instance 
study the impact of \textit{different fine-tuning} of the same encoder and of  encoders other than T5 and BART's. 
 Another direction would be to look at \textit{different parts of the model}, i.e. not only the output of the encoder but also the different layers both in the encoder and in the decoder.\\
Studying how NLG models perform regarding omissions and distortions when trained on \textit{different RDF-to-Text datasets} could also provide valuable insights on the causes for these mistakes.\\
Other analysis methods than probing can also be used. Indeed, an intrinsic limitation of probing is that it relies on correlations. Using \textit{causal methods}, for example by performing interventions on the embeddings themselves, would be an important step for further understanding omissions in pretrained language models.

\paragraph{Subjectivity in Manual Annotation}
For the annotation task, we do not formally define a threshold between omissions and distortions; rather, we let the annotators decide what they consider as omitted and distorted. The following examples show entities that annotators annotated differently in similar contexts.

\small{\textbf{Example (a):} The texts 1 and 2 are generated from a tripleset containing the triple Alan\_Shepard | selectedByNasa | 1959. In Text 1 the annotator considered the entity "1959" as mentioned, whereas the annotator of Text 2 considered it as distorted.
    
    \textit{Text 1: Alan Shepard was an American born in New Hampshire. He graduated from NWC with an M.A. in 1957 and was selected by NASA the same year. He died in California.}
    
    \textit{Text 2: American Alan Shepard was born in New Hampshire on November 18th, 1923. He graduated from NWC with an M.A. in 1957 and was hired by NASA the same year. He served as a test pilot and died in California.}}

\small{\textbf{Example (b):} 
    The annotator of Text 1 considered the entity 'GMA\_Network\_Center' to be omitted, whereas the annotator of Text 2 considered it as distorted.

    \textit{Text 1: GMA New Media, Inc. (parent company GMA Network) is an entertainment industry. Philippine Entertainment Portal and Digify Inc. are just two of the company’s subsidiaries.}
    
    \textit{Text 2: GMA New Media, Inc. (parent company GMA Network) is located in the Philippines. The company is involved in the entertainment industry with Philippine Entertainment Portal and Digify Inc. It is located inside GMA Center.}}

\paragraph{Application to an improved omissions detection}
While our probing methods help assess whether omissions can be tracked back to the encoder, \cite{faille-etal-2021-entity-based}'s algorithm permits detecting omissions based on a model-agnostic comparison between input graph and generated texts. These two methods could be combined to help facilitate the analysis of omissions in arbitrary RDF-to-Text generation models as follows. First, the omission detection algorithm can be applied to the output of the generation model to automatically annotate omissions. Second, these annotations can be used either as is or after manual validation, to fine-tune our probing classifier on the internal representations of the model under consideration. More generally, our methods help provide fine-grained information about both the quantity of omissions generated by a given model and the degree to which these omissions can be assumed to come from the encoder.  


\section{Conclusion}
This work took off from the hypothesis that omissions and distortions in KG-to-Text NLG models are due in large measure to issues with the way input entities are encoded.
We collected the first dataset for the analysis of omissions and distortions in the output of BART-based RDF-to-Text model and introduced two probes for understanding which parts of the model internals such semantic errors can be associated with. In addition to a standard, parametric, probing classifier, we introduced a parameter-free method which is based on the similarity between the embeddings of graphs associated with omission or distortions vs. the embeddings of graphs corrupted by removing either the omitted/distorted entity or an entity correctly mentioned in the output text. Both methods support the hypothesis that the encoding of graphs associated with omissions and distortions differs from the encoding of graphs which are not associated with such errors. Thus, we conclude, in line with our hypothesis, that the encoder plays an important role in the omission and/or distortion of input elements.

\section*{Acknowledgments}

We would like to thank the reviewers and the action editor for their insightful comments and suggestions which helped us greatly improve this paper. We would also like to thank Vadim Fomin for his very useful advice on probing classifiers.

Research reported in this paper has received funding from the European Union’s Horizon 2020 research and innovation programme under the Marie Skłodowska-Curie grant agreement No 860621 and from the French National Research Agency (Gardent; award ANR-20-CHIA-0003, XNLG "Multi-lingual, Multi-Source Text Generation"). This document reflects the views  of  the  authors  and  does  not  necessarily reflect the views or policy of the European Commission. The REA cannot be held responsible for any use that may be made of the information this document contains.


\bibliography{tacl2021}
\bibliographystyle{acl_natbib}

\appendix

\section{RDF-to-Text Models details}
\label{appendix:bart_details}
Our fine-tuning experiments are done using 2 Nvidia GTX 1080 Ti GPUs (11 GiB memory). We use pretrained models from the HuggingFace library \url{https://huggingface.co/facebook/bart-large} and \url{https://huggingface.co/t5-small}.
\paragraph{BART}
We fine-tune the BART large model which has 12 encoder and decoder layers and 400M parameters. We use the AdamW optimizer with an initial learning rate of $1e^{-5}$. We train during 12 epochs on WebNLG Train, using as input linearised RDF graphs without special tokens. For the decoding, we use greedy decoding with a maximum length of the generated sequence of 100 tokens and without length penalty. 
We evaluate the model on the WebNLG test set and obtain a corpus BLEU score of 0.31 \citep{papineni-etal-2002-bleu}, a chrF score of 0.53 \citep{popovic-2015-chrf} and Bert score precision, recall and f-score of 0.86 \citep{https://doi.org/10.48550/arxiv.1904.09675}. 

We find that these results are lower than those of the submissions of the WebNLG Challenge using BART \citep{yang-etal-2020-improving-text, montella-etal-2020-denoising} and pretraining on some external datasets, such as DocRED \citep{yao2019DocRED} or 103 millions of sentences extracted from Wikipedia. However they are comparable to the submission's result from \citet{sobrevilla-cabezudo-pardo-2020-nilc-webnlg} which did not use any additional pretraining.

\section{Subsidiary experiment on parametric probing: Omissions vs Distortions}
\label{appendix:omissions_distortions}
Table \ref{fig:trained_or_tested_on_distortions_omissions} shows the results of the neural probes trained on the subset with only omissions (Manual-O), and tested on the subset with only distortions  (Manual-D) and inversely, trained on the subset with only distortions  (Manual-D) and tested on the subset with only omissions  (Manual-O). The drop in results compared to training and testing on the same dataset shows that the probe trained on omissions does not generalise well to detecting distortions and vice versa. This is in line with the results of the parameter-free probing described in Section~\ref{subsection:similarity_results} and  suggests that omissions and distortions are errors of a different nature -- they can both be identified in the encoder output but seem to have qualitatively different representations.

\begin{table}[!htbp]
  \centering\small
  \begin{tabular}{ll|rr}
      &  &   \multicolumn{2}{c}{\bf Training Data}\\
      & & Manual-O & Manual-D  \\ \hline
    & \bf Test Data &&\\
    \multirow{2}{*}{1 Layer} &Manual-O & 0.60 (0.72) & 0.26 (0.5) \\
     &Manual-D & 0.19 (0.49) & 0.71 (0.79) \\
     \multirow{2}{*}{2 Layers} &Manual-O & 0.69 (0.77) & 0.27 (0.51) \\
           &Manual-D & 0.22 (0.48) & 0.79 (0.79) \\
    \end{tabular}
  \caption{F-measure of class 0 (Balanced accuracy) when training the neural classifiers on Omissions and testing on Distortions and vice versa}
   \label{fig:trained_or_tested_on_distortions_omissions}
\end{table}

\paragraph{T5}
We fine-tune a T5 small model 6 layers in encoder and decoder and 60M parameters. We use the Adafactor optimizer with $2e^{-5}$ learning rate. We use early stopping and trained for 72 epochs, with a 16 batch size.
We evaluate the model on the WebNLG test set and obtain a corpus BLEU score of 0.34, a chrF score of 0.58 and Bert score precision, recall and f-score of 0.88.

\subsection{Hyperparameters Tuning}
\label{sec:appendix_hyperparameter_tuning}
We tune hyperparameters of the neural classifiers trained on Manual-O, Manual-D, and on Manual-O+D. We use grid search with batch size (8, 16, 32, 64, 128, 256), initial learning rate (0.1, 0.01, 0.001, 0.0001) and size of the second layer (1000, 500, 100, 50, 10) for the 2 layer classifier. We choose the classifier that maximizes the f-measure of class 1.
In the table \ref{tab:hyperparameters}, we report the hyperparameters and the f-measure on the validation set of the best classifiers.

\begin{table}[!htbp]
\small
\begin{tabular}{l|llll}
Model & Batch Size & Size L2 & Lr   & F1 \\\hline
 N1, HP-O & 128 & - & 0.001 & 0.91     \\
 N1, HP-O+D  & 64 & - & 0.001 & 0.90 \\
 N1, HP-D & 256 & - & 0.01 & 0.91 \\ 
 N2, HP-O& 128 & 1000 & 0.01 & 0.93     \\
 N2, HP-O+D & 16 & 100 & 0.001 & 0.93 \\ 
 N2, HP-D & 32 & 1000 & 0.001 & 0.93 \\ 
 \multicolumn{5}{l}{\footnotesize HP:hyperparameters, O:omissions, D:distortions}\\
 \multicolumn{5}{l}{\footnotesize Lr:initial learning rate, L2:second layer of the classifier}
\end{tabular}
\caption{Hyperparameters of the best classifiers}
\label{tab:hyperparameters}
\end{table}

\section{Description of the annotation task}
\label{sec:appendix_annotation_instructions}
The annotation task took a total of 587 hours.\\
The following instructions were given to the annotators:\\
\textit{An automatically generated text is given, as well as input entities. For each input entity, please choose between 3 options:}
\begin{itemize}[noitemsep,nolistsep]
\item \textit{Mentioned: the entity is mentioned in the text}
\item \textit{Not Mentioned: the entity is not in the text}
\item \textit{Distorted: the entity is not in the text, but some “distorted” version of it is in the text 
}
\end{itemize}

\paragraph{Notes} {\small

\begin{itemize}[noitemsep,nolistsep]
\item \textit{
Please, try not to pay attention to the lack of fluency, grammar or meaning of some texts. What is important for us is whether an entity is present in a text or not, independently of the logic or quality of the text.}
\item \textit{Variants or Referring Expressions of an entity are considered as mentions. In other words, an entity doesn’t have to be expressed in the exact same way as in the input data, e.g.
125800 mm / 125 m, 
1703-05-27 / May 27th 1703,
Appleton,Winsconsin / Appleton.}
\item \textit{Disambiguation information of entities is sometimes given inside parentheses, like for example in The\_Honeymoon\_Killers\_(American\_band) or Federal\_Assembly\_(Switzerland). This information doesn’t necessarily have to be in the text.} 
\item \textit{Missing accents, missing spaces and capitalization problems are small mistakes and can be considered as mentions, e.g. “Cremazie station” can be considered as a mention of the entity “Crémazie station”}
\end{itemize}

\begin{table}[!htbp]
\centering
\small
\begin{tabularx}{0.95\columnwidth}{X|X}
\textbf{RDF Entity}                                       & \textbf{Distortion}                            \\ \hline
Olga\_Bondareva                                  & Olgaondarev                           \\
177539.0                                         & 1777539                               \\
Ciudad\_Ayala                                    & Ciudad Ayalatus                       \\
Lee Jae-hak                                      & Lee Lee-hak                           \\
Doosan Bears                                     & Donosan Bears                         \\
Lionsgate                                        & Lionsburg                             \\
1997                                             & 1996                                  \\
EGBF                                             & EAWFB                                 \\
Columbia\_Records                                & The Columbus Records                  \\
1929-06-11                                       & June 5th, 1929                        \\
St.\_Louis,\_Missouri                            & St Louis, Mississippi                 \\
11.51147.0                                       & 11.5                                  \\
-6                                               & Delta 6                              
\end{tabularx}
\caption{Examples of distortions found in Manual-D. The first column shows input RDF entities and the second column, their distortion in the output text. 
}
\label{fig:examples_distortions}
\end{table}
    
\section{Logistic regression features}
\label{appendix:features_logistic_regression}
For the logistic regression, we use 
the following input features which we can split into three main categories:

(1) Features about the tripleset: number of triples in the tripleset, category of the tripleset (in WebNLG: e.g. ‘Athlete’, ‘Scientist’, ’Food’..., in KELM: ‘’), position of the first occurrence of the entity in the tripleset, number of occurrences of the entity in the tripleset, nature of the entity in the tripleset (i.e. agent, patient or bridge as defined by \citet{castro-ferreira-etal-2018-enriching})}

(2) Features about the entity: DBPedia entity type, number of characters, whether the entity is a date, whether the entity is a number, entity shape (uppercase, lowercase, digits, other), entity shape (vowels, consonants)
    as used in \citet{van-der-goot-etal-2018-bleaching}

(3) Feature about the dataset: frequency of the entity in the training set of WebNLG

\end{document}